\documentclass[11pt,a4paper]{article}
\usepackage[hyperref]{emnlp2018}
\usepackage{times}
\usepackage{latexsym}
\usepackage{url}
\usepackage{graphicx}
\usepackage{algorithmic}
\usepackage{algorithm}
\usepackage{amsmath, amsfonts}
\usepackage{multirow}
\usepackage{booktabs}
\usepackage{tikz-dependency}

\aclfinalcopy 
\def\aclpaperid{1052} 

\newcommand{\argmax}{\mathop{\rm argmax}\limits}

\title{A Span Selection Model for Semantic Role Labeling}

\author{Hiroki Ouchi$^{2,3}$ \hspace{0.7cm} Hiroyuki Shindo$^{1,2}$ \hspace{0.7cm} Yuji Matsumoto$^{1,2}$\\
  {$^1$ Nara Institute of Science and Technology}\\
  {$^2$ RIKEN Center for Advanced Intelligence Project (AIP)}\\
  {$^3$ Tohoku University}\\
  {\tt hiroki.ouchi@riken.jp}, \{ {\tt shindo, matsu \}@is.naist.jp}}

\date{}

\begin{document}
\maketitle
\begin{abstract}
We present a simple and accurate span-based model for semantic role labeling (SRL).
Our model directly takes into account all possible argument spans and scores them for each label.
At decoding time, we greedily select higher scoring labeled spans.
One advantage of our model is to allow us to design and use span-level features, that are difficult to use in token-based BIO tagging approaches.
Experimental results demonstrate that our ensemble model achieves the state-of-the-art results, 87.4 F1 and 87.0 F1 on the CoNLL-2005 and 2012 datasets, respectively.
\end{abstract}

\section{Introduction}
\label{sec:intro}
Semantic Role Labeling (SRL) is a shallow semantic parsing task whose goal is to recognize the predicate-argument structure of each predicate.
Given a sentence and a target predicate, SRL systems have to predict semantic arguments of the predicate.
Each argument is a {\it span}, a unit that consists of one or more words.
A key to the argument span prediction is how to represent and model spans.

One popular approach to it is based on BIO tagging schemes.
State-of-the-art neural SRL models adopt this approach \cite{zhou:15,he:17,tan:18}.
Using features induced by neural networks, they predict a BIO tag for each word.
Words at the beginning and inside of argument spans have the ``B" and ``I" tags, and words outside argument spans have the tag ``O."
While yielding high accuracies, this approach reconstructs argument spans from the predicted BIO tags instead of directly predicting the spans.

Another approach is based on labeled span prediction \cite{oscar:15,fitzgerald:15}.
This approach scores each span with its label.
One advantage of this approach is to allow us to design and use span-level features, that are difficult to use in BIO tagging approaches.
However, the performance has lagged behind that of the state-of-the-art BIO-based neural models.

To fill this gap, this paper presents a simple and accurate span-based model.
Inspired by recent span-based models in syntactic parsing and coreference resolution \cite{stern:17,lee:17},  our model directly scores all possible labeled spans based on span representations induced from neural networks.
At decoding time, we greedily select higher scoring labeled spans.
The model parameters are learned by optimizing log-likelihood of correct labeled spans.

We evaluate the performance of our span-based model on the CoNLL-2005 and 2012 datasets \cite{carreras:05,pradhan:12}.
Experimental results show that the span-based model outperforms the BiLSTM-CRF model.
In addition, by using contextualized word representations, ELMo \cite{peters:18}, our ensemble model achieves the state-of-the-art results, 87.4 F1 and 87.0 F1 on the CoNLL-2005 and 2012 datasets, respectively.
Empirical analysis on these results shows that the label prediction ability of our span-based model is better than that of the CRF-based model.
Another finding is that ELMo improves the model performance for span boundary identification.

In summary, our main contributions include:
\begin{itemize}
\setlength{\parskip}{0cm} 
\setlength{\itemsep}{0cm} 
\item A simple span-based model that achieves the state-of-the-art results.
\item Quantitative and qualitative analysis on strengths and weaknesses of the span-based model.
\item Empirical analysis on the performance gains by ELMo.
\end{itemize}

\noindent
Our code and scripts are publicly available.\footnote{https://github.com/hiroki13/span-based-srl}

\section{Model}
\label{sec:problem}
We treat SRL as {\it span selection}, in which we select appropriate spans from a set of possible spans for each label.
This section formalizes the problem and provides our span selection model.

\subsection{Span Selection Problem}
\subsection*{Problem Setting}
Given a sentence that consists of $T$ words $w_{1:T} = w_1, \cdots, w_T$ and the target predicate position index $p$, the goal is to predict a set of labeled spans $Y = \{ \langle i, j, r \rangle_k \}^{|Y|}_{k=1}$.
\begin{align*}
\textbf{Input} \: & \text{:} \: X = \{ w_{1:T}, p \}, \\
\textbf{Output} \: & \text{:} \: Y = \{ \langle i, j, r \rangle_k \}^{|Y|}_{k=1} \:\:.
\end{align*}

\noindent
Each labeled span $\langle i, j, r \rangle$ consists of word indices $i$ and $j$  in the sentence ($1 \le i \le j \le T$) and a semantic role label $r \in \mathcal{R}$.

One simple method to predict $Y$ is to select the highest scoring span $(i, j)$ from all possible spans $\mathcal{S}$ for each label $r$,
\begin{align}
\label{eq:argmax}
\argmax_{(i, j) \in \mathcal{S}} \: \textsc{Score}_r(i, j), \: r \in \mathcal{R} \:\: .
\end{align}

\noindent
Function $\textsc{Score}_r(i, j)$ returns a real value for each span $(i, j) \in \mathcal{S}$ (described in Section \ref{sec:score} in more detail).
The number of possible spans $\mathcal{S}$ in the input sentence $w_{1:T}$ is $\frac{T (T +1)}{2}$, and $\mathcal{S}$ is defined as follows,
\[
\mathcal{S} = \{ (i, j) \: | \: i, j \in \{1, \cdots, T\}, i \le j \} \:\: .
\]

\noindent
Note that some semantic roles may not appear in the sentence.
To deal with the absence of some labels, we define the predicate position span $(p, p)$ as a {\sc Null} span and train a model to select the {\sc Null} span when there is no span for the label.\footnote{Since the predicate itself can never be an argument of its own, we define the position as the {\sc Null} span.}

\subsection*{Example}
Consider the following sentence with the set of correct labeled spans $Y$.\\

\hspace{1.5cm} She$_1$ \hspace{0.2cm} \underline{kept}$_2$ \hspace{0.2cm} a$_3$ \hspace{0.2cm} cat$_4$ 

\hspace{1.4cm} [\hspace{0.05cm} {\tt A0} \hspace{0.05cm}] \hspace{1.1cm} [ \hspace{0.3cm} {\tt A1} \hspace{0.3cm}]
\begin{align*}
Y = \{ \: & \langle 1, 1, \texttt{A0} \rangle, \: \langle 3, 4, \texttt{A1} \rangle,\\
& \langle 2, 2, \texttt{A2} \rangle, \cdots, \langle 2, 2, \texttt{TMP} \rangle \: \}
\end{align*}

\noindent
The input sentence is $w_{1:4} = \text{``She kept a cat"}$, and the target predicate position is $p=2$.
The correct labeled span $\langle 1, 1, \texttt{A0} \rangle$ indicates that the A0 argument is ``She", and $\langle 3, 4, \texttt{A1} \rangle$ indicates that the A1 argument is ``a cat".
The other labeled spans $\langle 2, 2, * \rangle$ indicate there are no arguments.

All the possible spans in this sentence are as follows,
\begin{align*}
\mathcal{S}_{w_{1:4}} = \{ & (1, 1), (1, 2), (1, 3), (1, 4), (2, 2), \\
& (2, 3), (2, 4), (3, 3), (3, 4), (4, 4) \} \:\: ,
\end{align*}

\noindent
where the predicate span $(2, 2)$ is treated as the \textsc{Null} span.
Among these candidates, we select the highest scoring span for each label.
As a result, we can obtain correct labeled spans $Y$.

\subsection{Scoring Function}
\label{sec:score}
As the scoring function for each span in Eq.~\ref{eq:argmax}, we model normalized distribution over all possible spans $\mathcal{S}$ for each label $r$,
\begin{align}
\label{eq:score}
\textsc{Score}_r(i, j) & = \text{P}_{\theta}(i, j \: | \: r) \nonumber \\
& = \frac{\text{exp}(\text{F}_{\theta}(i, j, r))}{\displaystyle \sum_{(i', j') \in \mathcal{S}} \text{exp}(\text{F}_{\theta}(i', j', r))} \:\:,
\end{align}

\noindent
where function $\text{F}_{\theta}$ returns a real value.

We train the parameters $\theta$ of $\text{F}_{\theta}$ on a training set,
\begin{align*}
\mathcal{D} & = \{ (X^{(n)}, Y^{(n)})\}_{n=1}^{|\mathcal{D}|} \:\:, \\
X & = \{ w_{1:T}, p \} \:\:, \\
Y & = \{ \langle i, j, r \rangle_k \}_{k=1}^{|Y|} \:\:.
\end{align*}

\noindent
To train the parameters $\theta$ of $\text{F}_{\theta}$, we minimize the cross-entropy loss function,

\begin{align}
\label{eq:loss}
\mathcal{L}({\theta}) & = \sum_{(X, Y) \in \mathcal{D}} \ell_{\theta}(X, Y) \:\:, \\ \nonumber
\ell_{\theta}(X, Y) & = \sum_{\langle i, j, r\rangle \in Y} \text{log} \: \text{P}_{\theta}(i, j | r) \:\:,
\end{align}

\noindent
where function $\ell_{\theta}(X, Y)$ is a loss for each sample.

\subsection{Function $\text{F}_{\theta}$}
\label{sec:function}
Function $\text{F}_{\theta}$ in Eq.~\ref{eq:score} consists of three types of functions; the base feature function $f_{base}$, the span feature function $f_{span}$ and the labeling function $f_{label}$ as follows,
\begin{align}
\label{a} {\bf h}_{1:T} & = f_{base}(w_{1:T}, p) \:\:, \\
\label{b} {\bf h}_s & = f_{span}({\bf h}_{1:T}, s) \:\:, \\
\label{c} \text{F}_{\theta}(i, j, r) & = f_{label}({\bf h}_s, r) \:\:.
\end{align}

\noindent
Firstly, $f_{base}$ calculates a base feature vector ${\bf h}_t$ for each word $w_t \in w_{1:T}$.
Then, from a sequence of the base feature vectors ${\bf h}_{1:T}$, $f_{span}$ calculates a span feature vector ${\bf h}_s$ for a span $s = (i, j)$.
Finally, using ${\bf h}_s$, $f_{label}$ calculates the score for the span $s = (i, j)$ with a label $r$.

Each function in Eqs.~\ref{a}, \ref{b} and \ref{c} can arbitrarily be defined.
In Section \ref{sec:network}, we describe our functions used in this paper.

\subsection{Inference}
\label{sec:inference}

The simple argmax inference (Eq.~\ref{eq:argmax}) selects one span for each label.
While this argmax inference is computationally efficient, it faces the following two problematic issues.
\begin{description}
\setlength{\parskip}{0cm} 
\setlength{\itemsep}{0cm} 
\item[(a)] The argmax inference sometimes selects spans that overlap with each other.
\item[(b)] The argmax inference cannot select multiple spans for one label.
\end{description}

\noindent
In terms of (a), for example, when $\langle 1, 3, \texttt{A0} \rangle$ and $\langle 2, 4, \texttt{A1} \rangle$ are selected, a part of these two spans overlaps.
In terms of (b), consider the following sentence.\\\vspace{-0.2cm}

\hspace{0.1cm} He \underline{came} to the U.S. yesterday at 5 p.m.

[{\tt A0}] \hspace{0.75cm} [\hspace{0.35cm} {\tt A4} \hspace{0.35cm}]\hspace{0.1cm}[\hspace{0.2cm} {\tt TMP} \hspace{0.2cm}]\hspace{0.1cm}[\hspace{0.1cm} {\tt TMP} \hspace{0.1cm}]\\

\vspace{-0.2cm}
\noindent
In this example, the label {\tt TMP} is assigned to the two spans (``yesterday" and ``at 5 p.m.").
Semantic role labels are mainly categorized into (i) {\it core labels} or (ii) {\it adjunct labels}.
In the above example, the labels {\tt A0} and {\tt A4} are regarded as core labels, which indicate obligatory arguments for the predicate.
In contrast, the labels like {\tt TMP} are regarded as adjunct labels, which indicate optional arguments for the predicate.
As the example shows, adjunct labels can be assigned to multiple spans.

To deal with these issues, we use a greedy search that keeps the consistency among spans and can return multiple spans for adjunct labels.
Specifically, we greedily select higher scoring labeled spans subject to two constraints.
\begin{description}
\setlength{\parskip}{0cm} 
\setlength{\itemsep}{0cm} 
\item[Overlap Constraint:] Any spans that overlap with the selected spans cannot be selected.
\item[Number Constraint:] While multiple spans can be selected for each adjunct label, at most one span can be selected for each core label.
\end{description}

\noindent
As a precise description of this algorithm, we describe the pseudo code and its explanation in Appendix \ref{sec:decode}. 

\section{Network Architecture}
\label{sec:network}
To compute the score for each span, we have introduced three functions ($f_{base}, f_{span}, f_{label}$) in Section \ref{sec:function}.
As an instantiation of each function, we use neural networks.
This section describes our neural networks for each function and the overall network architecture.

\subsection{BiLSTM-Span Model}
\label{sec:bilstm}
\begin{figure}[t]
  \begin{center}
   \includegraphics[width=7.5cm]{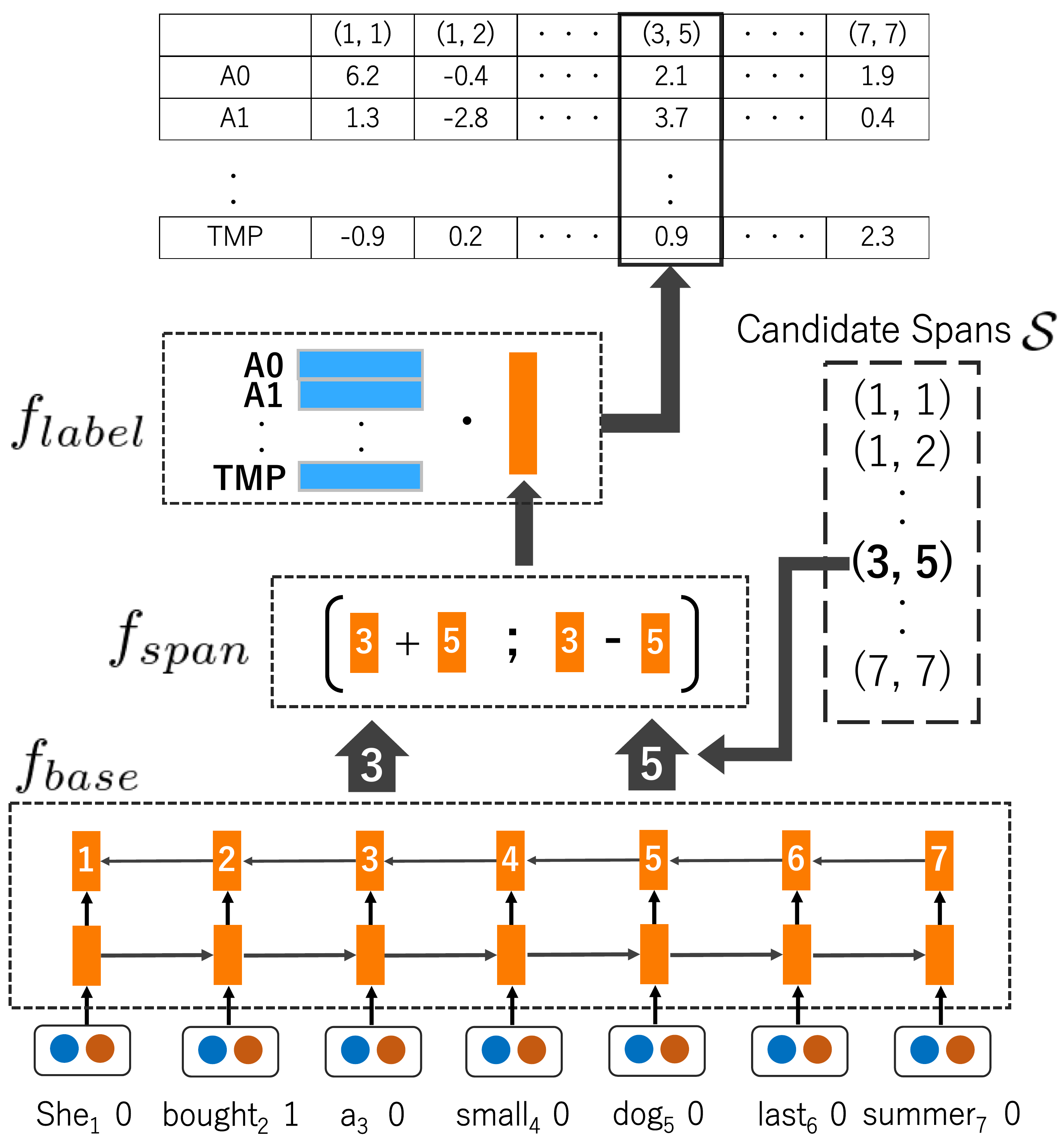}
  \caption{Overall architecture of our BiLSTM-span model.}
  \label{fig:model}
  \end{center}
\end{figure}

Figure \ref{fig:model} illustrates the overall architecture of our model.
The first component $f_{base}$ uses bidirectional LSTMs (BiLSTMs) \cite{schuster:97,graves:05,graves:13} to calculate the base features.
From the base features, the second component $f_{span}$ extracts span features.
Based on them, the final component $f_{label}$ calculates the score for each labeled span.
In the following, we describe these three components in detail. 

\subsection*{Base Feature Function}
As the base feature function $f_{base}$, we use BiLSTMs,
\begin{align*}
f_{base}(w_{1:T}, p) = \textsc{BiLSTM}(w_{1:T}, p) \:\: .
\end{align*}

\noindent
There are some variants of BiLSTMs.
Following the deep SRL models proposed by \newcite{zhou:15} and \newcite{he:17}, we stack BiLSTMs in an interleaving fashion.
The stacked BiLSTMs process an input sequence in a left-to-right manner at odd-numbered layers and in a right-to-left manner at even-numbered layers.

The first layer of the stacked BiLSTMs receives word embeddings ${\bf x}^{word} \in \mathbb{R}^{d^{word}}$ and predicate mark embeddings ${\bf x}^{mark} \in \mathbb{R}^{d^{mark}}$.
As the word embeddings, we can use existing word embeddings.
The mark embeddings are created from the mark feature which has a binary value.
The value is 1 if the word is the target predicate and 0 otherwise.
For example, at the bottom part of Figure~\ref{fig:model}, the word ``bought" is the target predicate and assigned $1$ as its mark feature.

Receiving these inputs, the stacked BiLSTMs calculates the hidden states until the top layer.
We use these hidden states as the input feature vectors ${\bf h}_{1:T}$ for the span feature function $f_{span}$ (Eq.~\ref{b}).
Each vector ${\bf h}_t \in {\bf h}_{1:T}$ has $d^{hidden}$ dimensions.
We provide a detailed description of the stacked BiLSTMs in Appendix~\ref{sec:lstm}.

\subsection*{Span Feature Function}
From the base features induced by the BiLSTMs, we create the span feature representations,
\begin{align}
\label{eq:span}
f_{span}({\bf h}_{1:T}, s) = [ {\bf h}_i + {\bf h}_j; {\bf h}_i - {\bf h}_j] \:\:,
\end{align}

\noindent
where the addition and subtraction features of the $i$-th and $j$-th hidden states are concatenated and used as the feature for a span $s = (i, j)$.
The resulting vector ${\bf h}_s$ is a $2d^{hidden}$ dimensional vector.

The middle part of Figure~\ref{fig:model} shows an example of this process.
For the span $(3, 5)$, the span feature function $f_{span}$ receives the $3$rd and $5$th features (${\bf h}_3$ and ${\bf h}_5$).
Then, these two vectors are added, and the $5$th vector is subtracted from the $3$rd vector.
The resulting vectors are concatenated and given to the labeling function $f_{label}$.

Our design of the span features is inspired by the span (or segment) features used in syntactic parsing \cite{wang:16,stern:17,teranishi:17}.
While these neural span features cannot be used in BIO-based SRL models, they can easily be incorporated into span-based models.

\subsection*{Labeling Function}
Taking a span representation ${\bf h}_s$ as input, the labeling function $f_{label}$ returns the score for the span $s = (i, j)$ with a label $r$.
Specifically, we use the following labeling function,
\begin{align}
\label{eq:label}
f_{label}({\bf h}_s, r) = {\bf W}[r] \cdot {\bf h}_s \:\:,
\end{align}

\noindent       
where ${\bf W} \in \mathbb{R}^{|\mathcal{R}| \times 2d^{hidden}}$ has a row vector associated with each label $r$, and ${\bf W}[r]$ denotes the $r$-th row vector.
As the result of the inner product of ${\bf W}[r]$ and ${\bf h}_s$, we obtain the score for a span $(i, j)$ with a label $r$.

The upper part of Figure~\ref{fig:model} shows an example of this process.
The span representation ${\bf h}_s$ for the span $s = (3, 5)$ is created from addition and subtraction of ${\bf h}_3$ and ${\bf h}_5$.
Then, we calculate the inner product of ${\bf h}_s$ and ${\bf W}[r]$.
The score for the label {\tt A0} is $2.1$, and the score for the label {\tt A1} is $3.7$.
In the same manner, by calculating the scores for all the spans $\mathcal{S}$ and labels $\mathcal{R}$, we can obtain the score matrix (at the top part of Figure~\ref{fig:model}).

\subsection{Ensembling}
\label{sec:mos}
We propose an ensemble model that uses span representations from multiple models.
Each base model trained with different random initializations has variance in span representations.
To take advantage of it, we introduce a variant of a mixture of experts (MoE) \cite{shazeer:17},
\footnote{One popular ensemble model for SRL is the product of experts (PoE) model \cite{fitzgerald:15,he:17,tan:18}.
In our preliminary experiments, we tried the PoE model but it did not improve the performance.}
\begin{align}
\label{eq:e1}
{\bf h}^{\text{moe}}_s & = {\bf W}^{\text{moe}}_s \cdot \sum^{M}_{m=1} \alpha_m \: {\bf h}^{(m)}_s \:\:,\\
\label{eq:e2}
f^\text{moe}_{label}({\bf h}^\text{moe}_s, r) & = {\bf W}^{\text{moe}}[r] \cdot {\bf h}^{\text{moe}}_s\:\:.
\end{align}

\noindent
Firstly, we combine span representations ${\bf h}^{(m)}_s$ from each model $m \in \{1, \cdots, M \}$.
${\bf W}^\text{moe}_s$ is a parameter matrix and $\{ \alpha_m \}^M_{m=1}$ are trainable, softmax-normalized parameters.
Then, using the combined span representation ${\bf h}^{\text{moe}}_s$, we calculate the score in the same way as Eq.~\ref{eq:label}.
We use the same greedy search algorithm used for our base model (Section~\ref{sec:inference}).

During training, we update only the parameters of the ensemble model, i.e., $\{ {\bf W}^{\text{moe}}_s, {\bf W}^{\text{moe}}, \{ \alpha_m \}^M_{m=1} \}$.
That is, we fix the parameters of each trained model $m$.
As the loss function, we use the cross-entropy (Eq.~\ref{eq:loss}).

\section{Experiments}
\label{sec:exp}

\subsection{Datasets}
We use the CoNLL-2005 and 2012 datasets\footnote{We use the version of OntoNotes downloaded at: http://cemantix.org/data/ontonotes.html.}.
We follow the standard train-development-test split and use the official evaluation script\footnote{The script can be downloaded at: http://www.lsi.upc.edu/~srlconll/soft.html} from the CoNLL-2005 shared task on both datasets.

\begin{table*}[t]
  \centering
  {\small
  \begin{tabular}{llcccccccccccc} \toprule
                  & & \multicolumn{3}{c}{Development}
                  & \multicolumn{3}{c}{Test WSJ}
                  & \multicolumn{3}{c}{Test Brown}
                  & \multicolumn{3}{c}{Test ALL} \\
                  {\sc Emb} & {\sc Model} & P & R & F1 & P & R & F1 & P & R & F1 & P & R & F1 \\ \hline
\multirow{3}{*}{\sc Senna}  & {\sc Crf}     & 81.7	& 81.3	& 81.5 & 83.3	& 82.5 &	82.9 & 72.6 &	70.0 &	71.3 & 81.9	& 80.8	& 81.4 \\
& {\sc Span}       & 83.6	& 81.4	& 82.5 & 84.7	& 82.3 &	83.5 & 76.0 	& 70.4 &	73.1& 83.6	& 80.7	& 82.1 \\
& {\sc Span} (Ensemble)      & 85.6	& 82.6	& 84.1 & 86.6	& 83.6 &	85.1 & 78.2 	& 71.8 &	74.8& 85.5	& 82.0	& 83.7 \\ \hline
\multirow{3}{*}{\sc ELMo}  & {\sc Crf} & 86.6	& 86.8	& 86.7 & 87.4	& 87.3	& 87.3 & 78.5	& 78.3	& 78.4 & 86.2	& 86.1	& 86.1\\
& {\sc Span}       & 87.4	& 86.3	& 86.9 & 88.2	& 87.0	& 87.6 & 79.9 &	77.5 &	78.7 & 87.1	& 85.7	& 86.4 \\
& {\sc Span} (Ensemble)       & {\bf 88.0}	& {\bf 86.9}	& {\bf 87.4} & {\bf 89.2}	& {\bf 87.9}	& {\bf 88.5} & {\bf 81.0} &	{\bf 78.4} &	{\bf 79.6} & {\bf 88.1}	& {\bf 86.6}	& {\bf 87.4} \\ \toprule
  \end{tabular}
  }
  \caption{\label{tab:result:conll05} Experimental results on the CoNLL-2005 dataset, in terms of precision (P), recall (R) and F1. The bold numbers denote the highest precision, recall and F1 scores among all the models.}
\end{table*}

\begin{table*}[t]
  \centering
  {\small
  \begin{tabular}{llcccccc} \toprule
                  & & \multicolumn{3}{c}{Development}
                  & \multicolumn{3}{c}{Test} \\
                  {\sc Emb} & {\sc Model} & P & R & F1 & P & R & F1 \\ \hline
                  \multirow{3}{*}{\sc Senna}  & {\sc Crf}     & 82.8 & 81.9 & 82.4 & 82.9 & 81.9 & 82.4 \\
& {\sc Span}       & 84.3 & 81.5 & 82.9 & 84.4 & 81.7 & 83.0 \\
& {\sc Span} (Ensemble)      & 86.0 & 83.0 & 84.5 & 86.1 & 83.3 & 84.7 \\ \hline
\multirow{3}{*}{\sc ELMo}  & {\sc Crf} & 86.1 &	{\bf 85.8} &	85.9 & 86.0	 & {\bf 85.7}	& 85.9\\
& {\sc Span}      & 87.2 &	85.5 &	86.3 & 87.1	& 85.3	& 86.2 \\
& {\sc Span} (Ensemble)       & {\bf 88.6} &	85.7 &	{\bf 87.1} & {\bf 88.5}	& 85.5	& {\bf 87.0} \\ \toprule
  \end{tabular}
  }
  \caption{\label{tab:result:conll12} Experimental results on the CoNLL-2012 dataset.}
\end{table*}

\begin{table}[t]
  \centering
  {\small
  \begin{tabular}{lcccc} \toprule
  		   & \multicolumn{3}{c}{CoNLL-05} & CoNLL12 \\
                  & WSJ & Brown & ALL & \\ \hline \hline
                  \multicolumn{5}{c}{\sc Single Model} \\ \hline
{\sc ELMo-Span}       & {\bf 87.6} & 78.7 & {\bf 86.4} & {\bf 86.2} \\
He+ 18 & 87.4 & {\bf 80.4} & - & 85.5 \\
Peters+ 18   & - & - & - & 84.6 \\
Strubell+ 18 & 83.9 & 72.6 & - & - \\
Tan+ 18   & 84.8 & 74.1 & 83.4 & 82.7 \\ 
He+ 17     & 83.1 & 72.1 & 81.6 & 81.7 \\ 
Zhou+ 15  & 82.8 & 69.4 & 81.1 & 81.3 \\ 
FitzGerald+ 15  & 79.4 & 71.2 & - & 79.6 \\
T\"{a}ckstr\"{o}m+ 15 & 79.9 & 71.3 & - & 79.4 \\
Toutanova+ 08 & 79.7 & 67.8 &  - & - \\ 
Punyakanok+ 08 & 79.4 & 67.8 & 77.9 & -\\ \toprule
                  \multicolumn{5}{c}{\sc Ensemble Model} \\ \hline
{\sc ELMo-Span}       & {\bf 88.5} & {\bf 79.6} & {\bf 87.4} & {\bf 87.0} \\
Tan+ 18   & 86.1 & 74.8 & 84.6 & 83.9 \\ 
He+ 17 & 84.6 & 73.6 & 83.2 & 83.4 \\ 
FitzGerald+ 15  & 80.3 & 72.2 & - & 80.1 \\
Toutanova+ 08 & 80.3 & 68.8 &  - & -\\ 
Punyakanok+ 08 & 79.4 & 67.8 & 77.9 & -\\ \toprule
  \end{tabular}
  }
  \caption{\label{tab:result:comparison} Comparison with existing models. The numbers denote F1 scores on each test set.}
\end{table}

\subsection{Baseline Model}
For comparison, as a model based on BIO tagging approaches, we use the BiLSTM-CRF model proposed by \newcite{zhou:15}.
The BiLSTMs for the base feature function $f_{base}$ are the same as those used in our BiLSTM-span model.
\vspace{-0.1cm}

\subsection{Model Setup}
\label{sec:model-setup}
As the base function $f_{base}$, we use 4 BiLSTM layers with 300 dimensional hidden units.
To optimize the model parameters, we use Adam \cite{kingma:14}.
Other hyperparameters are described in Appendix~\ref{sec:hparam} in detail.

\subsection*{\bf Word Embeddings}
Word embeddings have a great influence on SRL models.
To validate the model performance, we use two types of word embeddings.
\begin{itemize}
\setlength{\parskip}{0cm} 
\setlength{\itemsep}{0cm} 
\item Typical word embeddings, SENNA\footnote{http://ronan.collobert.com/senna/} \cite{collobert:11}
\item Contextualized word embeddings, ELMo\footnote{http://allennlp.org/elmo} \cite{peters:18}
\end{itemize}

\noindent
SENNA and ELMo can be regarded as different types of embeddings in terms of the context sensitivity.
SENNA and other typical word embeddings always assign an identical vector to each word regardless of the input context.
In contrast, ELMo assigns different vectors to each word depending on the input context.
In this work, we use these word embeddings that have different properties.\footnote{In our preliminary experiments, we also used the GloVe embeddings \cite{pennington:14}, but the performance was worse than SENNA.}
These embeddings are fixed during training.

\subsection*{Training}
As the objective function, we use the cross-entropy $\mathcal{L}_{\theta}$ in Eq.~\ref{eq:loss} with L2 weight decay,
\begin{equation}
\label{eq:reg}
\mathcal{L}_{\theta} = \sum_{(X, Y) \in \mathcal{D}} \ell_{\theta}(X, Y)+ \frac{\lambda}{2} ||\theta||^2 \:\:,
\end{equation}

\noindent
where the hyperparameter $\lambda$ is the coefficient governing the L2 weight decay.

\subsection{Results}
\label{sec:label}
We report averaged scores across five different runs of the model training.

Tables~\ref{tab:result:conll05} and \ref{tab:result:conll12} show the experimental results on the CoNLL-2005 and 2012 datasets.
Overall, our span-based ensemble model using ELMo achieved the best F1 scores, 87.4 F1 and 87.0 F1 on the CoNLL-2005 and CoNLL-2012 datasets, respectively.
In comparison with the CRF-based single model, our span-based single model consistently yielded better F1 scores regardless of the word embeddings, {\sc Senna} and {\sc ELMo}.
Although the performance difference was small between these models using {\sc ELMo}, it seems natural because both models got much better results and approached to the performance upper bound.

Table~\ref{tab:result:comparison} shows the comparison with existing models in F1 scores.
Our single and ensemble models using {\sc ELMo} achieved the best F1 scores on all the test sets except the Brown test set.

\section{Analysis}
\label{sec:analysis}

To better understand our span-based model, we addressed the following questions and obtained the following findings.

\subsection*{Questions}
\begin{description}
\setlength{\parskip}{0cm} 
\setlength{\itemsep}{0cm} 
\item[(a)] What are strengths and weaknesses of our span-based model compared with the CRF-based model?
\item[(b)] What aspect of SRL does ELMo improve?
\end{description}

\subsection*{Findings}
\begin{description}
\setlength{\parskip}{0cm} 
\setlength{\itemsep}{0cm} 
\item[(a)] While the CRF-based model is better at span boundary identification (Section~\ref{sec:sbi}), the span-based model is better at label prediction, especially for A2 (Section~\ref{sec:lp}).
 \item[(b)] ELMo improves the model performance for span boundary identification (Section~\ref{sec:sbi}).
\end{description}

\noindent
In addition, we have conducted qualitative analysis on span and label representations learned in the span-based model (Section~\ref{sec:qa}).

\subsection{Performance for Span Boundary Identification}
\label{sec:sbi}
\begin{table}[t]
  \centering
  {\small
  \begin{tabular}{llcccc} \toprule
                  & & \multicolumn{2}{c}{CoNLL-05} & \multicolumn{2}{c}{CoNLL-12} \\
                  {\sc Emb}& {\sc Model} & F1 & diff & F1 & diff \\ \hline
\multirow{2}{*}{\sc Senna} & {\sc Span}    & 86.6 & \multirow{2}{*}{-0.4}& 87.3 & \multirow{2}{*}{-0.6} \\
 & {\sc Crf}       & 87.0 &  & 87.9 & \\ \hline
\multirow{2}{*}{\sc ELMo} & {\sc Span}     & 90.5 & \multirow{2}{*}{-0.7} & 90.3 &\multirow{2}{*}{-0.6}  \\
 & {\sc Crf}        & 91.2 &   & 90.9 &  \\ \toprule
  \end{tabular}
  }
  \caption{\label{tab:span-identification} F1 scores only for span boundary match.}
\end{table}

We analyze the results predicted  by the single models.
We evaluate F1 scores only for the span boundary match, shown by Table~\ref{tab:span-identification}.
We regard a predicted boundary $\langle i, j, * \rangle$ as correct if it matches the gold annotation regardless of its label.

On both datasets, the CRF-based models achieved better F1 than that of the span-based models.
Also, compared with {\sc Senna}, {\sc ELMo} yielded much better F1 by over 3.0.
This suggests that a factor of the overall SRL performance gain by {\sc ELMo} is the improvement of the model ability to identify span boundaries.

\subsection{Performance for Label Prediction}
\label{sec:lp}
\begin{table}[t]
  \centering
  {\small
  \begin{tabular}{llcccc} \toprule
                  & & \multicolumn{2}{c}{CoNLL-05} & \multicolumn{2}{c}{CoNLL-12} \\
                  {\sc Emb}& {\sc Model} & Acc. & diff & Acc. & diff \\ \hline
\multirow{2}{*}{\sc Senna} & {\sc Span}    & 95.3 & \multirow{2}{*}{+1.5} & 95.1 & \multirow{2}{*}{+1.5} \\
 & {\sc Crf}       & 93.8 &  & 93.6 & \\ \hline
\multirow{2}{*}{\sc ELMo} & {\sc Span}     & 96.1 & \multirow{2}{*}{+0.9}  & 95.7 & \multirow{2}{*}{+1.3} \\
 & {\sc Crf}        & 95.2 &   & 94.4 &  \\ \toprule
  \end{tabular}
  }
  \caption{\label{tab:label-prediction} Accuracies only for semantic role labels.}
\end{table}

We analyze labels of the predicted results.
For labeled spans whose boundaries match the gold annotation, we evaluate the label accuracies.
As Table~\ref{tab:label-prediction} shows, the span-based models outperformed the CRF-based models.
Also, interestingly, the performance gap between {\sc Senna} and {\sc ELMo} was not so big as that for span boundary identification.

\subsection*{Label-wise Performance}
\begin{table*}[t]
  \centering
  {\small
  \begin{tabular}{l|rrrr|rrrr} \toprule
   & \multicolumn{4}{c}{CoNLL-2005} & \multicolumn{4}{c}{CoNLL-2012} \\
   & \multicolumn{2}{c}{\sc Senna} & \multicolumn{2}{c}{\sc ELMo} & \multicolumn{2}{c}{\sc Senna} & \multicolumn{2}{c}{\sc ELMo} \\
   Label & \multicolumn{1}{c}{\sc Crf} & \multicolumn{1}{c}{\sc Span} & \multicolumn{1}{c}{\sc Crf} & \multicolumn{1}{c}{\sc Span} & \multicolumn{1}{c}{\sc Crf} & \multicolumn{1}{c}{\sc Span} & \multicolumn{1}{c}{\sc Crf} & \multicolumn{1}{c}{\sc Span} \\ \hline
   A0   & 89.9 & 90.2 &  93.0 & 93.2 & 89.9 & 90.0 & 92.5 & 92.5\\
   A1   & 83.2 & 83.8 &  89.1& 89.2  & 84.7 & 85.1 & 88.7 & 89.0\\
   A2   & 70.9 &  73.1 &  80.0 & 81.2 & 78.6 & 79.4 & 83.2 & 84.2\\
   A3   & 64.4 &  71.2 &  78.8 & 78.5 & 61.9 & 62.9 & 69.0 & 70.7\\
   ADV & 59.3 & 61.9 & 68.1 & 67.0 & 63.2 & 63.7 & 67.5 & 67.0\\
   DIR & 43.2 &  47.3 &  56.6& 54.5 & 54.1 & 52.0 & 61.1 & 59.7\\
   LOC        & 58.2 &  60.5 &  68.1& 68.3 & 65.8 & 65.0 & 72.0 & 72.0\\
   MNR & 61.4 & 61.3 &  66.5& 67.7 & 64.4 &  65.7& 70.5 & 71.1\\
   PNC & 57.3 & 60.2 &  68.8& 67.7 & 18.5 & 13.7 & 20.2 & 16.1\\
   TMP & 81.8 & 82.7 & 86.1 & 86.0 & 82.2 & 82.3 & 86.1 & 86.2\\ \hline
   Overall & 81.5 & 82.5 & 86.7 & 86.9 & 82.4 & 82.9 & 85.9 & 86.3\\ \toprule
  \end{tabular}
  }
  \caption{\label{tab:label:result} F1 Scores for frequent labels on the development set of the CoNLL-2005 and 2012 datasets.}
\end{table*}

Table~\ref{tab:label:result} shows F1 scores for frequent labels on the CoNLL-2005 and 2012 datasets.
For A0 and A1, the performances of the CRF-based and span-based models were almost the same.
For A2, the span-based models outperformed the CRF-based model by about 1.0 F1 on the both datasets.
\footnote{The PNC label got low scores on the CoNLL-2012 dataset in Table\ \ref{tab:label:result}. Almost all the gold PNC (purpose) labels are assigned to only the news article domain texts of the CoNLL-2012 dataset. The other 6 domain texts have no or very few PNC labels. This can lead to the low performance.}

\subsection*{Label Confusion Matrix}
\begin{figure}[t]
  \begin{center}
    \includegraphics[width=7cm]{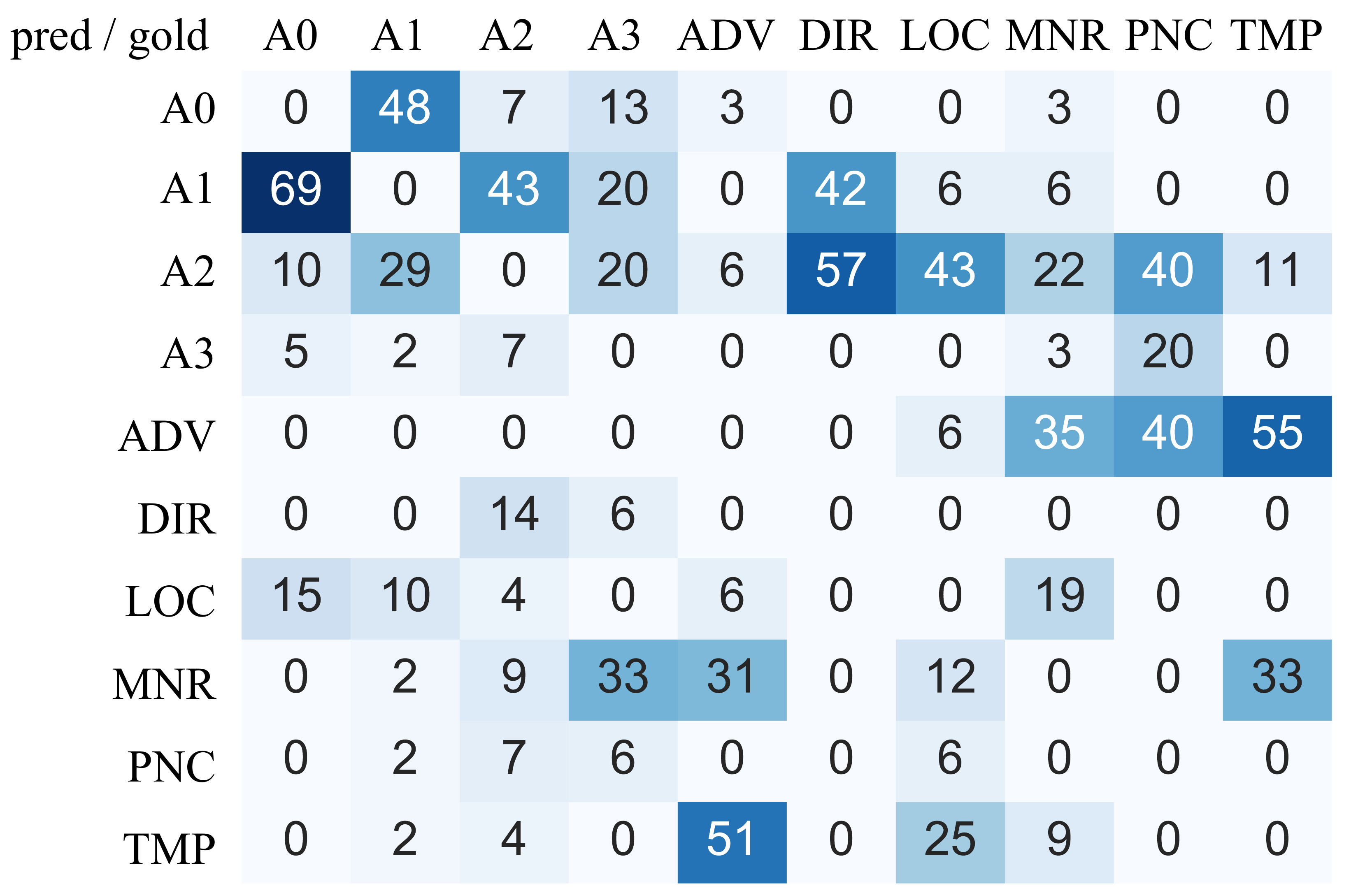}
  \end{center}
  \vspace{-0.2cm}
  \caption{\label{fig:label-conf} Confusion matrix for labeling errors of our span-based model using ELMo. Each cell shows the percentage of predicted labels for each gold label.}
\end{figure}

\noindent
Figure~\ref{fig:label-conf} shows a confusion matrix for labeling errors of the span-based model using ELMo.\footnote{We have observed the same tendency of labeling confusions between the models using ELMo and SENNA.}
Following \newcite{he:17}, we only count predicted arguments that match the gold span boundaries.

The span-based model confused A0 and A1 arguments the most.
In particular, the model confused them for ergative verbs.
Consider the following two sentences:\vspace{-0.3cm}\\

\noindent
\hspace{0.5cm} People \underline{start} their own business ... \\
\vspace{0.2cm}\hspace{0.5cm} [\hspace{0.1cm} {\tt A0} \hspace{0.1cm}] \\
\hspace{0.5cm} .. Congress has \underline{started} to jump on ... \\
\vspace{-0.3cm}\hspace{0.2cm} [\hspace{0.3cm} {\tt A1} \hspace{0.3cm}] \\

\noindent
where the constituents located at the syntactic subjective position fulfill a different role A0 or A1 according to their semantic properties, such as animacy.
Such arguments are difficult for SRL models to correctly identify.

Another point is the confusions of A2 with DIR and LOC.
As \newcite{he:17} pointed out, A2 in a lot of verb frames represents semantic relations such as direction or location, which can cause the confusions of A2 with such location-related adjuncts.
To remedy these two problematic issues, it can be a promising approach to incorporate frame knowledge into SRL models by using verb frame dictionaries.

\subsection{Qualitative Analysis on Our Model}
\label{sec:qa}

\subsection*{On Span Representations}
\begin{table}[t]
  \centering
  {\small
  \begin{tabular}{rcl} \toprule
  \multicolumn{3}{c}{``$\cdots$ toy makers to \underline{move} [ across the border ] ."} \\
  \multicolumn{3}{c}{\hspace{2.9cm}GOLD:A2} \\
  \multicolumn{3}{c}{\hspace{2.9cm}PRED:DIR} \\ \hline \hline
  \multicolumn{3}{c}{Nearest neighbors of ``across the border"}\\ \hline
1 & DIR & across the Hudson \\
2 & DIR & outside their traditional tony circle \\
3 & DIR & across the floor \\
4 & DIR & through this congress \\
5 & A2 & off their foundations \\
6 & DIR & off its foundation\\
7 & DIR & off the center field wall \\
8 & A3 & out of bed \\
9 & A2 & through cottage rooftops \\
10 & DIR & through San Francisco \\ \toprule
  \end{tabular}
  }
  \caption{\label{tab:knn} Example of the CoNLL-2005 development set, in which our model misclassified the label for the span ``across the border". We collect 10 nearest neighbors of this span from the training set.}
\end{table}

\noindent
Our span-based model computes and uses span representations (Eq.~\ref{eq:span}) for label prediction.
To investigate a relation between the span representations and predicted labels, we qualitatively analyze nearest neighbors of each span representation with its predicted label.
Specifically, for each predicted span in the development set, we collect 10 nearest neighbor spans with their gold labels from the training set.

Table~\ref{tab:knn} shows 10 nearest neighbors of a span ``across the border" for the predicate ``move".
The label of this span was misclassified, i.e., the predicted label is DIR but the gold is A2.
Looking at its nearest neighbor spans, they have different gold labels, such as DIR, A2 and A3.
Like this case, we have observed that spans with a misclassified label often have their nearest neighbors with inconsistent labels.

\subsection*{On Label Embeddings}
\begin{figure}[t]
  \begin{center}
    \includegraphics[width=7cm]{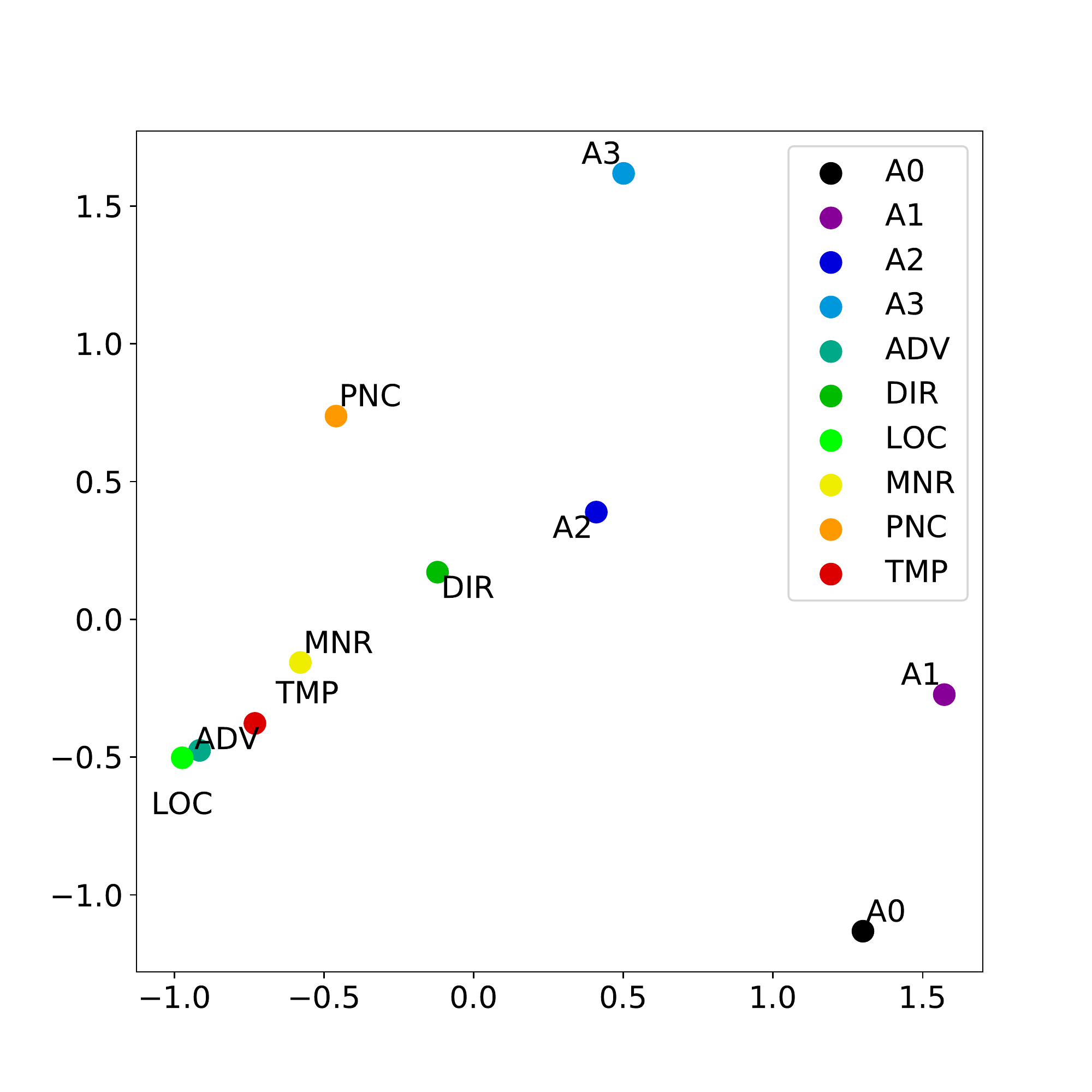}
  \end{center}
  \vspace{-0.8cm}
  \caption{\label{fig:label-emb} Label embedding distribution of our span-based model.}
\end{figure}

\noindent
We analyze the label embeddings in the labeling function (Eq.~\ref{eq:label}).
Figure~\ref{fig:label-emb} shows the distribution of the learned label embeddings.
The adjunct labels are close to each other, which are likely to be less discriminative.
Also, the core label A2 is close to the adjunct label DIR, which are often confused by the model.
To enhance the discriminative power, it is promising to apply techniques that keep label representations far away from each other \cite{wen:16,luo:17}.

\section{Related Work}
\label{sec:rwork}
\subsection{Semantic Role Labeling Tasks}

Automatic SRL has been widely studied \cite{gildea:02}.
There have been two main styles of SRL.
\begin{itemize}
\setlength{\parskip}{0cm} 
\setlength{\itemsep}{0cm} 
\item FrameNet-style SRL \cite{baker:98}
\item PropBank-style SRL \cite{palmer:05}
\end{itemize}

\noindent
In this paper, we have tackled PropBank-style SRL.\footnote{Detailed descriptions on FrameNet-style and PropBank-style SRL can be found in \newcite{baker:98,das:14,kingsbury:02,palmer:05}.}

In PropBank-style SRL, there have been two main task settings.
\begin{itemize}
\setlength{\parskip}{0cm} 
\setlength{\itemsep}{0cm} 
\item Span-based SRL: CoNLL-2004 and 2005 shared tasks \cite{carreras:04,carreras:05}
\item Dependency-based SRL:  CoNLL-2008 and 2009 shared tasks \cite{surdeanu:08,hajivc:09}
\end{itemize}

\begin{figure}[t]
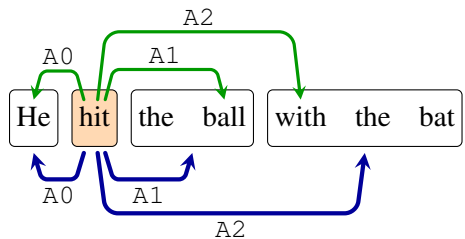

\centering
\begin{dependency}[text only label, label style={above}, edge style={green!60!black,very thick}]

\begin{deptext}[column sep=.2cm]
He \& hit \& the \& ball \& with \& the \& bat \\
\end{deptext}

\depedge[edge unit distance=0.38cm]{2}{1}{{\Large \tt A0}}
\depedge[edge unit distance=0.2cm]{2}{4}{{\Large \tt A1}}
\depedge[edge unit distance=0.3cm, edge start x offset=-0.1cm]{2}{5}{{\Large \tt A2}}

\wordgroup[minimum height=0.75cm]{1}{1}{1}{a0}
\wordgroup[minimum height=0.75cm]{1}{3}{4}{a1}
\wordgroup[minimum height=0.75cm]{1}{5}{7}{a2}
\wordgroup[minimum height=0.75cm, group style={fill=orange!30}]{1}{2}{2}{prd}

\groupedge[edge below, edge style={blue!60!black,ultra thick}, label style={below, yshift=-0.05cm}]{prd}{a0}{\Large \tt A0}{0.35cm}
\groupedge[edge below, edge style={blue!60!black,ultra thick}, label style={below, yshift=-0.05cm}]{prd}{a1}{\Large \tt A1}{0.35cm}
\groupedge[edge below, edge style={blue!60!black,ultra thick}, label style={below, yshift=-0.05cm}, edge start x offset=-0.1cm]{prd}{a2}{\Large \tt A2}{0.8cm}

\end{dependency}
\caption{\label{fig:dep:exam} Example of dependency-based SRL (the upper part) and span-based SRL (the lower part).}
\end{figure}

\noindent
Figure~\ref{fig:dep:exam} illustrates an example of span-based and dependency-based SRL.
In dependency-based SRL (at the upper part of Figure~\ref{fig:dep:exam}), the correct A2 argument for the predicate ``hit" is the word ``with".
On one hand, in span-based SRL (at the lower part of Figure~\ref{fig:dep:exam}), the correct A2 argument is the span ``with the bat".

For span-based SRL, the CoNLL-2004 and 2005 shared tasks \cite{carreras:04,carreras:05} provided the task settings and datasets.
In the task settings, various SRL models, from traditional pipeline models to recent neural ones, have been proposed and competed with each other \cite{pradhan:05,he:17,tan:18}.
For dependency-based SRL, the CoNLL-2008 and 2009 shared tasks \cite{surdeanu:08,hajivc:09} provided the task settings and datasets.
As in span-based SRL, recent neural models achieved high-performance in dependency-based SRL \cite{marcheggiani:17a,marcheggiani:17b,he:dep:18,cai:18}.
This paper focuses on span-based SRL.

\subsection{BIO-based SRL Models}

Span-based SRL can be solved as BIO sequential tagging \cite{hacioglu:04,pradhan:05,marquez:05}.\\

\noindent
{\bf Neural models} \hspace{0.2cm} State-of-the-art SRL models use neural networks based on the BIO tagging approach.
The pioneering neural SRL model was proposed by \newcite{collobert:11}.
They use convolutional neural networks (CNNs) and CRFs.
Instead of CNNs, \newcite{zhou:15} and \newcite{he:17} used stacked BiLSTMs and achieved strong performance without syntactic inputs.
\newcite{tan:18} replaced stacked BiLSTMs with self-attention architectures.
\newcite{strubell:18a} improved the self-attention SRL model by incorporating syntactic information.

\noindent
{\bf Word representations} \hspace{0.2cm} Typical word representations, such as SENNA \cite{collobert:11} and GloVe \cite{pennington:14}, have been used and contributed to the performance improvement \cite{collobert:11,zhou:15,he:17}.
Recently, \newcite{peters:18} integrated contextualized word representation, ELMo, into the model of \newcite{he:17} and improved the performance by 3.2 F1 score.
\newcite{strubell:18b} also integrated ELMo into the model of \newcite{strubell:18a} and reported the performance improvement.

\subsection{Span-based SRL Models}
Another line of approaches to SRL is labeled span modeling \cite{xue:04,koomen:05,toutanova:05}.\\

\vspace{-0.3cm}
\noindent
{\bf Typical models} \hspace{0.2cm} 
Typically, in this approach, models firstly identify candidate argument spans (argument identification) and then classify each span into one of the semantic role labels (argument classification).
For inference, several effective methods have been proposed, such as structural constraint inference by using integer linear programming \cite{punyakanok:08} or dynamic programming \cite{oscar:15,fitzgerald:15}.\\

\vspace{-0.3cm}
\noindent
{\bf Recent span-based model} \hspace{0.2cm} A very recent work, \newcite{he:18}, proposed a span-based SRL model similar to our model.
They also used BiLSTMs to induce span representations in an end-to-end fashion.
A main difference is that while they model $\text{P}(r | i, j)$, we model $\text{P}(i, j | r)$.
In other words, while their model seeks to select an appropriate label for each span ({\it label selection}), our model seeks to select appropriate spans for each label ({\it span selection}).
This point distinguishes between their model and ours.\\

\vspace{-0.3cm}
\noindent
{\bf FrameNet span-based model} \hspace{0.2cm} For FrameNet-style SRL, \newcite{swayamdipta:17} used a segmental RNN \cite{kong:16}, combining bidirectional RNNs with semi-Markov CRFs \cite{sarawagi:04}.
Their model computes span representations using BiLSTMs and learns a conditional distribution over all possible labeled spans of an input sequence.
Although we cannot compare our results with theirs, we can regard that our model is simpler and effective for PropBank-style SRL.

\subsection{Span-based Models in Other NLP Tasks}
In syntactic parsing, \newcite{wang:16} proposed an LSTM-based sentence segment embedding method named LSTM-Minus.
\newcite{stern:17,kitaev:18} incorporated the LSTM Minus into their parsing model and achieved the best results in constituency parsing.
In coreference resolution, \newcite{lee:17,lee:18} presented an end-to-end coreference resolution model, which considers all spans in a document as potential mentions and learn distributions over possible antecedents for each.
Our model can be regarded as an extension of their model.

\section{Conclusion and Future Work}
\label{sec:conc}
We have presented a simple and accurate span-based model.
We treat SRL as {\it span selection} and our model seeks to select appropriate spans for each label.
Experimental results have demonstrated that despite the simplicity, the model outperforms a strong BiLSTM-CRF model.
Also, our span-based ensemble model using ELMo achieves the state-of-the-art results on the CoNLL-2005 and 2012 datasets.
Through empirical analysis, we have obtained some interesting findings.
One of them is that the span-based model is better at label prediction compared with the CRF-based model.
Another one is that ELMo improves the model performance for span boundary identification.

An interesting direction for future work concerns evaluating span representations from our span-based model.
Since the investigation on the characteristics of the representations can lead to interesting findings, it is worthwhile evaluating them intrinsically and extrinsically.
Another promising direction is to explore methods of incorporating frame knowledge into SRL models.
We have observed that a lot of label confusions arise due to the lack of such knowledge.
The use of frame knowledge to reduce these confusions is a straightforward approach.

\section*{Acknowledgments}
This work was partially supported by JST CREST Grant Number JPMJCR1513 and JSPS KAKENHI Grant Number 18K18109.
We are grateful to the members of the NAIST Computational Linguistics Laboratory, the members of Tohoku University Inui-Suzuki Laboratory, Kentaro Inui, Jun Suzuki, Yuichiro Matsubayashi, and the anonymous reviewers for their insightful comments.

\bibliography{emnlp2018}
\bibliographystyle{acl_natbib_nourl}

\newpage
\appendix

\section{Span-Consistent Greedy Search}
\label{sec:decode}
\begin{algorithm}[H]
\caption{Span-Consistent Greedy Search}
\begin{algorithmic}[1]
\STATE {\bf Input:} Score Matrix ${\bf M} \in \mathbb{R}^{|\mathcal{R}| \times |\mathcal{S}|}$, 
\STATE \hspace{1.1cm} Predicate Position Index $p$
\STATE \hspace{1.1cm} Core Label Set $\mathcal{R}^{(\text{core})}$

\STATE spans $\leftarrow \phi$
\STATE used\_cores $\leftarrow \phi$

\STATE $\mathcal{U} \leftarrow \{ (i, j, r, \text{score}) \in flatten(\textbf{M}) \}$
\STATE $\mathcal{U} \leftarrow filter(\mathcal{U}, p)$
\STATE {\bf for} $(i, j, r, \text{score}) \in sort(\mathcal{U})$ {\bf do}
\STATE \hspace{0.25cm} {\bf if} $r \notin \text{used\_cores}$ {\bf and}
\STATE \hspace{0.6cm} $\text{is\_overlap}((i, j), \text{spans})$ {\bf is False} {\bf then}
\STATE \hspace{1cm} spans $\leftarrow$ spans $\cup \: \{ \langle i, j, r \rangle \}$
\STATE \hspace{1cm} {\bf if} $r \in \mathcal{R}^{(\text{core})}$ {\bf then}
\STATE \hspace{1.3cm} used\_cores $\leftarrow$ used\_cores $\cup \: \{ r \}$

\STATE {\bf return} spans
\end{algorithmic}
\end{algorithm}

\noindent
Algorithm 1 describes the pseudo code of the greedy search algorithm introduced in Section \ref{sec:inference}.
This algorithm receives the three inputs (line~1-3).
${\bf M}$ is the score matrix illustrated at the top part of Figure\ \ref{fig:model} in Section\ \ref{sec:network}.
Each cell of the matrix represents the score of each span.
$p$ is a target predicate position index.
$\mathcal{R}^\text{(core)}$ is the set of core labels.
At line~4, the variable ``spans" is initialized.
This variable stores the selected spans to be returned as the output.
At line~5, the variable ``used$\_$cores" is initialized.
This variable keeps track of the already selected core labels.

At line~6, the score matrix ${\bf M}$ is converted to tuples, $(i, j, r, \text{score})$, by the function $flatten(\cdot)$.
These tuples are stored in the variable $\mathcal{U}$.
At line~7, from $\mathcal{U}$, we remove the tuples that fall into any one of the followings, (i) the tuples whose boundary $(i, j)$ overlaps with the predicate position $p$ or (ii) the tuples whose score is lower than that of the predicate span tuples.
In terms of (i), since spans whose boundary $(i, j)$ overlaps with the predicate position, $i \leq p \leq j$, can never be a correct argument, we remove such tuples.
In terms of (ii), we remove the tuples $(*, *, r, \text{score})$ whose score is lower than that of the predicate span tuple $(p, p, r, \text{score})$.
In Section\ \ref{sec:problem}, we define the predicate span $(p, p)$ as the {\sc Null} span, implying that we can regard the spans whose score is lower than that of the {\sc Null} span as an inappropriate argument.
Thus, we remove such tuples from the set of the candidates $\mathcal{U}$.

The main processing starts from line~8.
Based on the scores, the function $sort(\cdot)$ sorts the tuples $(i, j, r, \text{score})$ in a descending order.
At line~9-10, there are constraints for output spans.
At line~9, ``$r \notin \text{used\_cores}$" represents the constraint that at most one span can be selected for each core label.
At line~10, the function $\text{is\_overlap}(\cdot)$ takes as input a span $(i, j)$ and the set of the selected spans, and returns the boolean value (``True" or ``False") that represents whether the span overlaps with any one of the selected spans or not.

At line~11, the span is added to the set of the selected spans.
At line~12-13, if the label $r$ is included in the core labels $\mathcal{R}^\text{(core)}$, the label is added to ``$\text{used\_cores}$".
At line~14, as the final output, the set of the selected spans ``$\text{spans}$" is returned.

\section{BiLSTMs}
\label{sec:lstm}

As the base feature function $f_{base}$ (Eq.~\ref{a} in Section~\ref{sec:function}), we use BiLSTMs,
\begin{align*}
f_{base}(w_{1:T}, p) = \textsc{BiLSTM}(w_{1:T}, p) \:\: .
\end{align*}

\noindent
In particular, we use the stacked BiLSTMs in an interleaving fashion \cite{zhou:15,he:17}.
The stacked BiLSTMs process an input sequence in a left-to-right manner for odd-numbered layers and in a right-to-left manner for even-numbered layers.

The stacked BiLSTMs consist of $L$ layers.
The hidden state in each layer $\ell \in \{1, \cdots , L\}$ is calculated as follows,
\begin{equation}
\nonumber
  {\bf h}^{(\ell)}_t = \begin{cases}
    \text{LSTM}^{(\ell)} ({\bf x}^{(\ell)}_t, \: {\bf h}^{(\ell)}_{t-1}) & (\ell \: \mbox{=} \: \mbox{odd}) \\
    \text{LSTM}^{(\ell)} ({\bf x}^{(\ell)}_t, \: {\bf h}^{(\ell)}_{t+1}) & (\ell \: \mbox{=} \: \mbox{even}) \:\: .
  \end{cases}
\end{equation}

\noindent
Both of the odd- and even-numbered layers receive ${\bf x}^{(\ell)}_t$ as the first input of the LSTM.
For the second input, odd-numbered layers receive ${\bf h}^{(\ell)}_{t-1}$, whereas even-numbered layers receive ${\bf h}^{(\ell)}_{t+1}$.

Between the LSTM layers, we use the following connection \cite{zhou:15},
\[
{\bf x}^{(\ell+1)}_t = \text{ReLU}({\bf W}^{(\ell)} \cdot [ {\bf x}^{(\ell)}_t ; {\bf h}^{(\ell)}_t ]) \:\:.
\]

\noindent
Here, we firstly concatenate ${\bf x}^{(\ell)}_t$ and ${\bf h}^{(\ell)}_t$, and then calculate the inner product of the concatenated vector and the parameter matrix ${\bf W}^{(\ell)}$ with the rectified linear units (ReLU).
As a result, we obtain the input representation ${\bf x}^{(\ell+1)}_t$ for the next ($\ell+1$-th) LSTM layer.

In the first layer, $\text{LSTM}^{(1)}$ receives an input feature vector ${\bf x}^{(1)}_t$.
Following \newcite{he:17}, we create this vector by concatenating a word embedding and predicate mark embedding,
\begin{align}
\nonumber
{\bf x}^{(1)}_t = [{\bf x}^{word}_t ; {\bf x}^{mark}_t] \:\:,
\end{align}

\noindent
where ${\bf x}^{word} \in \mathbb{R}^{d^{word}}$ and ${\bf x}^{mark} \in \mathbb{R}^{d^{mark}}$.
The mark embedding is created from the binary mark feature.
The value is 1 if the word is the target predicate and 0 otherwise.

After the $L$-th LSTM layer runs, we obtain ${\bf x}^{(L+1)}_{1:T} = {\bf x}^{(L+1)}_1, \cdots , {\bf x}^{(L+1)}_T$.
We use them as the input of the span feature function $f_{span}$ (Eq.~\ref{b} in Section~\ref{sec:function}), i.e., ${\bf h}_{1:T} = {\bf x}^{(L+1)}_{1:T}$.
Each vector ${\bf h}_t \in {\bf h}_{1:T}$ has $d^{hidden}$ dimensions.

\section{Hyperparameters}
\label{sec:hparam}
\begin{table}[H]
  \centering
  {\small
  \begin{tabular}{lr} \toprule
    Name & Value \\ \hline
    \multirow{2}{*}{Word Embedding $d^{word}$}
    & 50-dimensional SENNA \\
    & 1024-dimensional ELMo \\
    Mark Embedding $d^{mark}$ & 50-dimensional vector \\
    LSTM Layers $L$ & 4 \\
    LSTM Hidden Units $d^{hidden}$ & 300 dimensions \\
    Mini-batch Size & 32 \\
    Optimization & Adam \\
    Learning Rate & 0.001 \\
    L2 Regularization $\lambda$ & 0.0001 \\
    Dropout Ratio for BiLSTMs & 0.1 \\ 
    Dropout Ratio for ELMo & 0.5 \\ \toprule
  \end{tabular}
  }
  \caption{\label{tab:hyperparam} Hyperparameters for our span-based model.}
\end{table}

\subsection{Span-based Model}
Table \ref{tab:hyperparam} lists the hyperparameters used for our span-based model.\\

\vspace{-0.3cm}
\noindent
{\bf Word representation setup} \hspace{0.2cm} As word embeddings ${\bf x}^{word}$, we use two types of embeddings, (i) SENNA \cite{collobert:11}, 50-dimensional word vectors ($d^{word} = 50$), and (ii) ELMo \cite{peters:18}, 1024-dimensional vectors ($d^{word} = 1024$).
During training, we fix these word embeddings (not update them).
As predicate mark embeddings ${\bf x}^{mark}$, we use randomly initialized 50-dimensional vectors ($d^{mark} = 50$).
During training, we update them.\\

\vspace{-0.3cm}
\noindent
{\bf Network setup} \hspace{0.2cm} As the base feature function $f_{base}$, we use $4$ stacked BiLSTMs (2 forward and 2 backward LSTMs) with 300-dimensional hidden units ($d^{hidden} = 300$).
Following \newcite{he:17}, we initialize all the parameter matrices in BiLSTMs with random orthonormal matrices \cite{saxe:13}.
Other parameters are initialized following \newcite{glorot:10}, and bias parameters are initialized with zero vectors.

\noindent
{\bf Regularization}\hspace{0.2cm} We set the coefficient $\lambda$ for the L2 weight decay (Eq.~\ref{eq:reg} in Section~\ref{sec:model-setup}) to $0.0001$.
We apply dropout \cite{srivastava:14} to the input vectors of each LSTM with dropout ratio of 0.1 and the ELMo embeddings with dropout ratio of 0.5.\\

\noindent
{\bf Training} \hspace{0.2cm} To optimize the parameters, we use Adam \cite{kingma:14} with $\beta_1 = 0.9$ and $\beta_2 = 0.999$.
The learning rate is initialized to 0.001.
After training 50 epochs, we halve the learning rate every 25 epochs.
Parameter updates are performed in mini-batches of 32.
The number of training epochs is set to 100.
We save the parameters that achieve the best F1 score on the development set and evaluate them on the test set.
Training our model on the CoNLL-2005 training set takes about one day and on the CoNLL-2012 training set takes about two days on a single GPU, respectively.

\subsection{\bf Ensemble Model}
Our ensemble model uses span representations ${\bf h}^{(m)}_s$ from base models $m \in \{1, \cdots, M \}$ (Section \ref{sec:mos}).
We use 5 base models ($M = 5$) learned over different runs.
Note that, during training, we fix the parameters of the five base models and update only the parameters of the ensemble model.\\

\noindent
{\bf Network setup} \hspace{0.2cm} The parameter matrix ${\bf W}^\text{moe}_s$ (Eq.~\ref{eq:e1} in Section~\ref{sec:mos}) is initialized with the identity matrix.
The scalar parameters $\{ \alpha_m \}^M_{m=1}$ (Eq.~\ref{eq:e1}) are initialized with $0$.
Each row vector ${\bf W}^\text{moe}[r]$ of the parameter matrix ${\bf W}^\text{moe}$  (Eq.~\ref{eq:e2}) is initialized with the averaged vector over the row vectors ${\bf W}^{(m)}[r]$ of each model $m$, i.e., $\frac{1}{M} \sum_{m=1}^M {\bf W}^{(m)}[r]$.
\\

\noindent
{\bf Training} \hspace{0.2cm} To optimize the parameters, we use Adam with $\beta_1 = 0.9$ and $\beta_2 = 0.999$.
The learning rate is set to 0.0001.
Parameter updates are performed in mini-batches of 8.
The number of training epochs is set to 20.
We save the parameters that achieve the best F1 score on the development set and evaluate them on the test set.
Training one ensemble model on the CoNLL-2005 and 2012 training sets takes about one day on a single GPU.

\end{document}